\documentclass[journal]{IEEEtran}

\usepackage{xcolor, soul,framed} %,caption

\colorlet{shadecolor}{yellow}
\usepackage[pdftex]{graphicx}
\graphicspath{{../pdf/}{../jpeg/}}
\DeclareGraphicsExtensions{.pdf,.jpeg,.png}

\usepackage[cmex10]{amsmath}
\usepackage{array}
\usepackage{mdwtab}
\usepackage{eqparbox}
\usepackage{url}
\usepackage{multirow}
\usepackage{amssymb}

\usepackage{array}
\usepackage{graphicx}
\usepackage{hyperref}
\usepackage{bm}

\usepackage{tabularx}
\usepackage{caption}
\usepackage{footnote}
\usepackage{threeparttable}
\usepackage{booktabs}

\usepackage{changepage}

\usepackage{tcolorbox}

\begin{document}
\bstctlcite{IEEEexample:BSTcontrol}
    \title{VideoExpert: Augmented LLM for Temporal-Sensitive Video Understanding}
  %\author{
      %Henghao~Zhao,~\IEEEmembership{}
      % Erez Falkenstein,~\IEEEmembership{Student Member,~IEEE,}
      % and~Zechao~Li,~\IEEEmembership{Fellow,~IEEE}% <-this % stops a space
%}

  %\thanks{Henghao Zhao, Rui Yan, and Zechao Li are with the School of Computer Science and Engineering, Nanjing University of Science and Technology, Nanjing 210094, China. E-mail: \{henghaozhao, ruiyan, zechao.li\}@njust.edu.cn.}% <-this % stops a space
  % \thanks{E-mail: \{henghaozhao, ruiyan, zechao.li\}@njust.edu.cn.}%
  %\thanks{Corresponding author: Zechao Li.}% <-this % stops a space
        % }  
\author{Henghao Zhao,
        Ge-Peng Ji,
        Rui Yan,
        Huan Xiong
        and Zechao Li
        % <-this % stops a space

% \thanks{Manuscript received xx xx, xxxx; revised xx xx, xxxx.} % <-this % stops a space
\thanks{{H. Zhao, R. Yan and Z. Li  are with School of Computer Science and Engineering, Nanjing University of Science and Technology, Nanjing 210094, China.  E-mail: \{henghaozhao, ruiyan, zechao.li\}@njust.edu.cn.

G.P. Ji is with the School of Computing, Australian National University, Canberra 2601, Australia. E-mail: gepengai.ji@gmail.com.

H. Xiong is with the Institute for Advanced Study in Mathematics, Harbin Institute of Technology, Heilongjiang 150001, China. E-mail: huan.xiong.math@gmail.com.}

(Corresponding author: Huan Xiong and Zechao Li)
}}

\markboth{IEEE TRANSACTIONS ON xxxxx, VOL.~xx, NO.~xx, xx~2023
}{Roberg \MakeLowercase{\textit{et al.}}: VideoExpert: Augmented LLM for Temporal-sensitive Video Understanding}

% ====================================================================
\maketitle
% === ABSTRACT ====================================================================
% =================================================================================

%%==================================%%
%% abstract %%
%%==================================%%

\begin{abstract}

The core challenge in video understanding lies in perceiving dynamic content changes over time. However, multimodal  large language models (MLLMs) struggle with temporal-sensitive video tasks, such as video temporal grounding, which requires generating timestamps to mark the occurrence of specific events. Existing strategies require MLLMs to generate absolute or relative timestamps directly. We have observed that those MLLMs tend to rely more on language patterns than visual cues when generating timestamps, affecting their performance. To address this problem, we propose VideoExpert, a general-purpose MLLM suitable for several temporal-sensitive video tasks. Inspired by the expert concept, VideoExpert integrates two parallel modules: the Temporal Expert and the Spatial Expert. The Temporal Expert is responsible for modeling time sequences and performing temporal grounding. It processes high-frame-rate yet compressed tokens to capture dynamic variations in videos and includes a lightweight prediction head for precise event localization. The Spatial Expert focuses on content detail analysis and instruction following. It handles specially designed spatial tokens and language input, aiming to generate content-related responses. These two experts collaborate seamlessly via a special token \texttt{<LOC>}, ensuring coordinated temporal grounding and content generation. Notably, the Temporal and Spatial Experts maintain independent parameter sets. This parameter decoupling design enables specialized learning within each part without mutual interference. By offloading temporal grounding from content generation, VideoExpert prevents text pattern biases in timestamp predictions. Moreover, we introduce a Spatial Compress module to obtain spatial tokens. This module filters and compresses patch tokens while preserving key information, delivering compact yet detail-rich input for the Spatial Expert. Extensive experiments conducted on four widely-used benchmarks (i.e. Charades-STA, QVHighlight, YouCookII and NextGQA) across four tasks (temporal grounding, highlight detection, dense video captioning and grounding question answering) demonstrate the effectiveness and versatility of the VideoExpert.

\end{abstract}

% === KEYWORDS ====================================================================
% =================================================================================


% ====================================================================
% ====================================================================
% ====================================================================
% ====================================================================

% \fi
\IEEEpeerreviewmaketitle

% ====================================================================
% ====================================================================
% ====================================================================

% === I. INTRODUCTION =============================================================
% =================================================================================

%%%%%%%%%%%%%%%%%%%%%%
% Introduciton
%%%%%%%%%%%%%%%%%%%%%

\section{Introduction}

\IEEEPARstart{M}{ultimodal} large language models (MLLMs)~\cite{BLIP2, minigpt4, llava, videollama} offer a unique approach to video understanding by performing tasks like captioning and question answering, enabling us to interpret human knowledge across history and cultures within the visual stream from a new perspective. However, most top-performing models~\cite{videochat2, VideoChatGPT, videollava, TCSVT_videoLLM_1} focus primarily on overall content comprehension, lacking the ability to identify boundaries and dynamic relationships among events. This limitation hampers their performance in temporal-sensitive tasks, which require precise moment boundaries to pinpoint when specific events occur, such as temporal grounding, highlight detection, dense video captioning, and grounding question answering.

\begin{figure}[!tp]
  \centering
  \includegraphics[width=\linewidth]{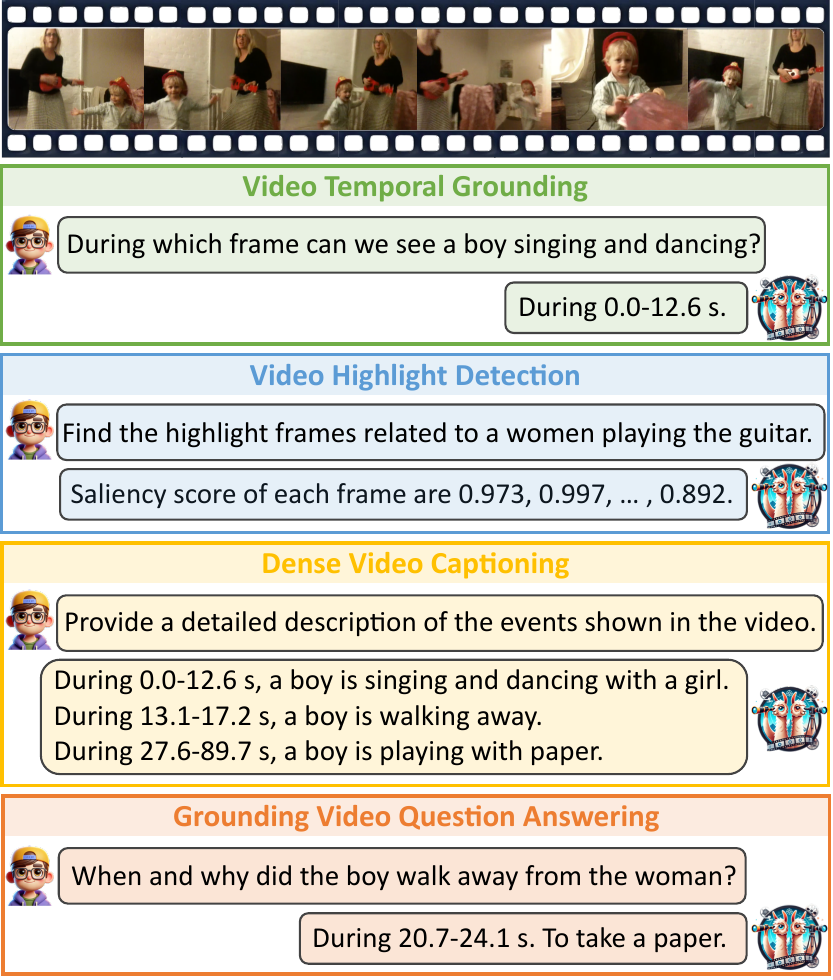}
  \caption{{An example illustrating the temporal-sensitive video understanding tasks addressed by VideoExpert.}}
  \label{fig_showcase}
  \vspace{-0.5 cm}
\end{figure}

\begin{figure*}[t]
  \begin{adjustwidth}{0 cm}{}
    \centering
    \includegraphics[width=1\linewidth]{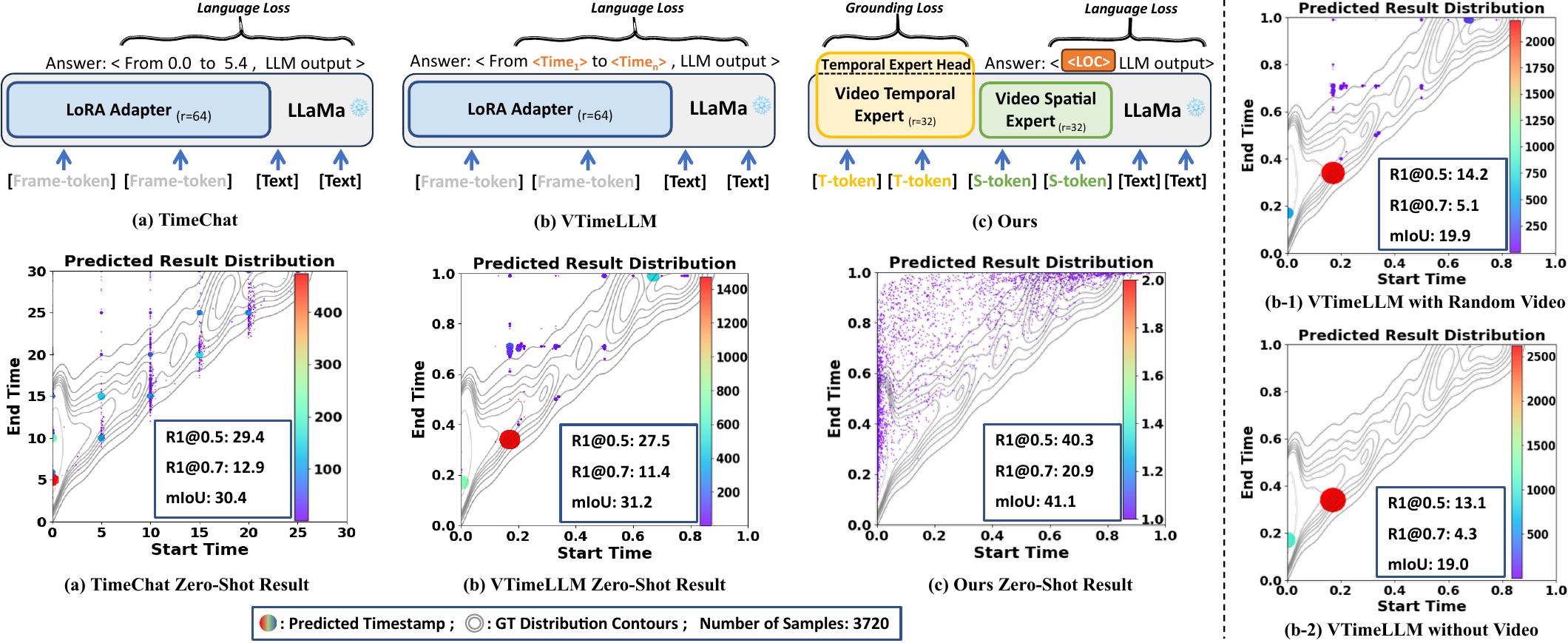}
  \end{adjustwidth}
  \caption{The predicted result distributions of TimeChat~\cite{timechat}, VTimeLLM~\cite{VTimeLLM} and our VideoExpert on the Charades-STA~\cite{twostage1} test split. Each point represents a predicted timestamp result. More prominent points indicate more frequent predictions. The contour plot shows the Ground-Truth distribution, where higher areas reflect more concentrated annotations. Both TimeChat and VTimeLLM frequently predict the same time range as result across different video-query pairs. This phenomenon becomes more pronounced when visual information is inaccurate or missing. These methods rely more on language patterns rather than visual cues when generating timestamps.}
  \label{fig_motivation}
  \vspace{-0.5 cm}
\end{figure*}

Existing MLLM methods that can be used for temporal-sensitive video tasks fall into three paradigms. The modality-switching paradigm~\cite{LLovi-modelchange, LangRePo-modelchange, omagent-modelchange, groundingprompter-modelchange, videoagent-modelchange} converts visual content into text via a captioning tool, which is then processed by a language model to analyze temporal relationships and content details. However, this approach often suffers from context loss and information omission due to the lack of direct visual perception. Its performance heavily depends on the quality of the captioning tool. The two-stage paradigm~\cite{sevila-twostage, videotree-twostage} identifies relevant clips through techniques such as frame scoring, followed by applying image-based MLLMs for response generation. While effective, this paradigm is prone to error accumulation, as the two stages operate independently. In contrast, the fine-tuning paradigm~\cite{timechat, LITA, VTimeLLM, Momentor,vtgllm} gains boundary perception capabilities by fine-tuning LLMs on video data. They convert temporal grounding into a text generation task by mapping timestamps to special tokens. The fine-tuning paradigm shows greater potential by jointly processing video and text input, avoiding the information loss caused by modality-switching. However, they still lag behind task-specific methods in performance.

The performance gap in the fine-tuning paradigm can mainly be attributed to the training mechanisms of LLMs. As probabilistic generation models, LLMs tend to predict the most ``common" pattern related to the generated context when performing next-token predictions. To illustrate this, we visualize the prediction results of two fine-tuning methods, TimeChat~\cite{timechat} and VTimeLLM~\cite{VTimeLLM}, as shown in Figure~\ref{fig_motivation}. These methods frequently predict similar time ranges as a result across different video-query pairs. When a video input is randomly selected or missing, visual cues fail to provide meaningful guidance for timestamp generation, and this phenomenon becomes more pronounced. Due to the use of fixed templates, the models may have memorized certain common time ranges as fixed expressions (i.e., the high appearance probability of time tokens) during fine-tuning, rather than learning to perform conditional time regression. As a result, these methods rely more on learned language patterns than on visual cues when generating timestamps. In addition, the performance of MLLMs is influenced by the quality of the visual input. To simulate temporal context, some methods use only class tokens as visual input, which are too abstract to provide sufficient information, particularly for fine-grained generation tasks. Although dense visual patches can offer richer information, they are inefficient due to redundancy between video frames. Therefore, the key challenge in improving model performance lies in reducing redundancy while maintaining sufficient visual details, striking a balance between information richness and computational efficiency.

To tackle these challenges, we propose VideoExpert, a multimodal language model suitable for various temporal-sensitive video understanding tasks. The core idea is to delegate temporal perception and content generation to specialized modules. Specifically, VideoExpert integrates two parallel components: the Temporal Expert and the Spatial Expert. The Temporal Expert processes high-frame-rate yet highly compressed feature input to capture dynamic information in videos. It focuses on modeling temporal relationships, such as event sequences and scene transitions. Additionally, a lightweight prediction head is incorporated to ensure accurate event localization. Meanwhile, the Spatial Expert specializes in capturing fine-grained content details and following instructions. By processing spatial tokens and language input, it generates content-related responses. During content generation, the Spatial Expert collaborates with the Temporal Expert via a special token, \texttt{<LOC>}, to indicate when and what events should be localized. By maintaining independent parameter sets, each expert encodes distinct types of information, enabling a collaborative framework that ensures a comprehensive understanding of video content. Importantly, the Spatial Expert only generates special tokens in place of explicit timestamps, delegating all temporal grounding requests to the Temporal Expert. This design eliminates reliance on text pattern biases in timestamp predictions. Furthermore, we introduce patch tokens to provide the model with richer fine-grained information. However, handling a large number of patch tokens presents a significant challenge. To address this, we implemented the Spatial Compress module. This module filters and compresses large-scale patch tokens while preserving key information, supplying the Spatial Expert with compact, detail-rich input.

To validate the VideoExpert, we conduct experiments not only on a temporal grounding benchmark (Charades-STA~\cite{twostage1}) but also on joint temporal grounding and highlight detection (QVHighlights~\cite{moment-detr}), dense video captioning (YouCookII~\cite{YouCookII}), and video grounding question answer (Next-GQA~\cite{NextGQA}) benchmarks. As a general MLLM model designed for temporal-sensitive video tasks, our method achieved remarkable results. It surpasses existing MLLM-based methods and even competing with task-specific models. The main contributions of this work are summarized as threefold:

\begin{itemize}

\item{A general-purpose MLLM is proposed in this paper, named VideoExpert, which is suitable for several temporal-sensitive video tasks.}

\item{VideoExpert integrates two parallel modules: the Temporal Expert for temporal perception and the Spatial Expert for content generation. Each expert focuses on specific task types, eliminating reliance on text pattern biases in timestamp predictions.}

\item{A Spatial Compress module is proposed to filter and compress large-scale patch tokens while preserving key information. This module supplies the Spatial Expert with compact, detail-rich inputs to enhance its detailed perception capability.}

\item{Extensive experiments conducted on four challenging datasets across four tasks, i.e. Charades-STA, QVHighlights, YouCookII and Next-GQA, demonstrate the effectiveness of the proposed method.}

\end{itemize}

% === II. Related Works ========================
% =================================================================================

%%%%%%%%%%%%%%%%%%%%%%
% Related Work
%%%%%%%%%%%%%%%%%%%%%

% \smallskip
\vspace{0.2 cm}
\section{Related Work}
This section reviews the progress of multimodal large language models and four temporal-sensitive video tasks.

\subsection{Multimodal Large Language Model}

In recent years, large language models (LLMs) have revolutionized natural language processing~\cite{llama, qwen2} and significantly impacted the field of computer vision~\cite{BLIP2, minigpt4, llava}. To enable LLMs to understand visual information, current approaches typically employ pre-trained image encoders (e.g., CLIP~\cite{CLIP}) to process visual inputs and map them into the textual embedding space via mechanisms such as Q-former~\cite{BLIP2} or linear projection~\cite{minigpt4, llava}, creating Multimodal LLMs (MLLMs). These studies have shown impressive capabilities on image-level tasks like captioning and question answering. However, They often struggle with region- or pixel-level fine-grained tasks. These problems have triggered another research trend. Researchers~\cite{lisa, pixelllm_lisa, GLaMM_lisa} have developed new methods that combine external expert modules to decode bounding boxes or masks. For example, LISA~\cite{lisa} leverages MLLMs to guide SAM~\cite{sam} in generating segmentation masks. While effective for fine-grained segmentation, this approach introduces additional latency during inference, as the MLLMs and SAM process the image independently.

A similar concept extends naturally from image to multi-frame video tasks. Most video-based LLM~\cite{videollama, videochat2, VideoChatGPT, videollava, minigpt4video} studies sample a few frames with large strides, prioritizing a holistic understanding of the video. However, this strategy is inadequate for fine-grained tasks, especially for time-sensitive applications such as video temporal grounding or dense video captioning. Existing methods for alleviating these limitations can be divided into three categories: 1) Language-based methods~\cite{LLovi-modelchange, LangRePo-modelchange, omagent-modelchange, groundingprompter-modelchange, videoagent-modelchange} that convert video content into text via pre-trained captioning models, which are subsequently processed by LLMs. However, the approaches lack visual perception capabilities, causing context loss and information omission. 2) Two-stage methods~\cite{sevila-twostage, videotree-twostage} that first identify relevant clips with techniques like frame-by-frame scoring, followed by applying MLLMs for further response generation. A drawback of this approach is the risk of compounding errors. 3) Direct Fine-Tuning methods~\cite{timechat, LITA, VTimeLLM, Momentor,vtgllm}, like TimeChat~\cite{timechat} and VTimeLLM~\cite{VTimeLLM}, which convert temporal grounding into a text generation task, and fine-tune the MLLMs end-to-end. However, a performance gap remains compared to traditional methods~\cite{moment-detr,vslnet}.

In contrast to previous approaches, the proposed VideoExpert integrates two experts to collaboratively process video inputs. The Temporal Expert is responsible for temporal perception and can directly performs content localization without converting the task into text generation. In conjunction with the LLM, the Spatial Expert focus on video detail and content generation. This division of labour minimizes mutual interference while facilitating effective multi-task collaboration. Moreover, VideoExpert processes the visual input only once, avoiding unnecessary latency.

\subsection{Temporal-sensitive Video Tasks}

\noindent
\textbf{Video Temporal Grounding} aims to locate specific moments in a video based on a text query. The methods in this task follows two paradigms: proposal-based and proposal-free. Proposal-based methods~\cite{twostage1, 2D-TAN, TCSVT_MR_proposal_1} rely on various proposal generation techniques and rank candidate proposals according to the query. In contrast, the proposal-free paradigm~\cite{proposalfree3, TNN_proposalfree_SDN} directly estimates the start and end boundaries of target moments without proposal candidates. A unique proposal-free approach, Moment-DETR~\cite{moment-detr}, treats the task as a set prediction problem, training the decoder to learn queries at different temporal scales to identify the relevant moments.

% The techniques include sliding windows, proposal generation networks~\cite{} and 2D maps~\cite{}.

\smallskip
\noindent \textbf{Video Highlight Detection} aims to identify engaging segments within a given video. The task is required to assign a saliency score to each video clip and select the highest-scoring clip as the result. Traditionally, datasets~\cite{tvsum, youtube} in this field are query-agnostic and lack the capability to tailor highlights according to specific queries. Lei et al.~\cite{moment-detr} introduced a new benchmark, QVHighlights, which enables users to customize video highlights based on their specific queries. They utilized the proposed Moment-DETR to assign the saliency scores. Subsequently, UMT~\cite{UMT} incorporated the audio modality to enrich the information, while QD-DETR~\cite{qddetr} introduced saliency tokens and developed negative pairs for contrastive learning. Overall, current methods~\cite{TCSVT_HD_1, TCSVT_HD_2} rely on ranking-based techniques, training models to assign higher scores to highlight clips using hinge loss, cross-entropy loss, contrastive loss, or reinforcement learning approaches.

\smallskip
\noindent \textbf{Dense Video Captioning} is a challenging task because it requires both event localization and captioning within the same framework. Traditional methods~\cite{dvc_twostage_2, dvc_twostage_3, dvc_twostage_1} often relied on a two-stage strategy, with separate phases for localization and captioning. Recent methods~\cite{dvc_onestage_3, dvc_onestage_2, dvc_onestage_1} emphasize improving task interaction by jointly training the localization and captioning modules. For example, Vid2Seq~\cite{vid2seq} incorporates specialized temporal tokens into LLMs, enabling the model to simultaneously generate event timestamps and textual descriptions in a single output sequence.

\smallskip
\noindent \textbf{Video Grounding Question Answering} requires models to understand video content to answer questions and to locate relevant segments as visual evidence. This process improves the reliability of answers and has applications in fields such as embodied vision~\cite{embodied_vision} and contextual memory enhancement~\cite{Episodic}. Like dense video captioning, VideoGQA has evolved from two-stage models~\cite{GQA_twostage_tvqa,GQA_twostage_tvqaplus, sevila-twostage} to more integrated joint learning approaches~\cite{ego_groundingQA}. Di et al.~\cite{ego_groundingQA} proposed an encoder-decoder model that uses an encoder to fuse video and question, with separate temporal and language decoders to predict event boundaries and generate answers, respectively.

% === III. Methods =======================================
% =================================================================================

%%%%%%%%%%%%%%%%%%%%%%
% Methods
%%%%%%%%%%%%%%%%%%%%%

\begin{figure*}[ht]
  \begin{adjustwidth}{0 cm}{}
    \centering
    \includegraphics[width=1\linewidth]{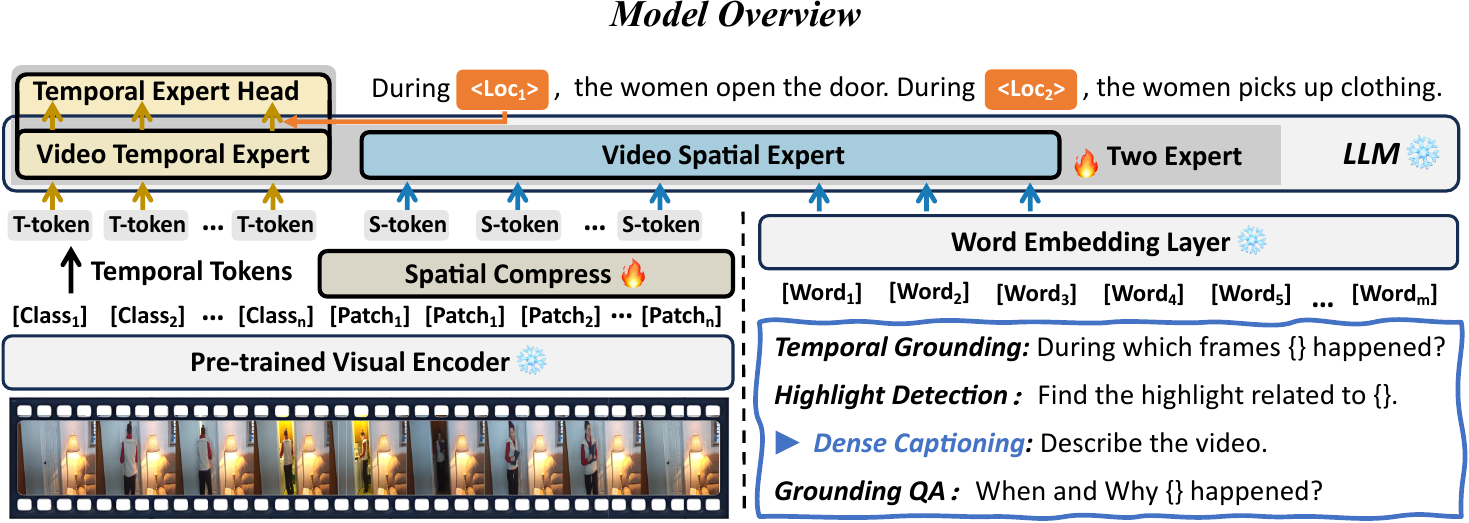}
  \end{adjustwidth}
  \caption{Overview of the proposed VideoExpert for a series of temporal-sensitive video tasks.}
  \label{fig_overview}
  \vspace{-0.2 cm}
\end{figure*}

\section{Methodology}

We propose VideoExpert, a model proficient in handling temporal-sensitive video tasks, as shown in Figure~\ref{fig_overview}. This section begins by providing an overview of the proposed approach. The Temporal and Spatial Expert component is detailed in Section B, while the Spatial Compress module is explained in Section C. The boundary-aware training paradigm is described in Section D. Finally, the training objective utilized in this model is presented in Section E.

\subsection{Model Overview}

VideoExpert, like most existing models, employs a pretrained encoder to process visual inputs and a visual-language adapter to map visual features into the language domain.The model then relies on a LLM to generate responses and complete tasks. The key innovation of our approach lies in the integration of two parallel expert modules within the LLM, which focus on temporal perception and content generation for video-related tasks. Additionally, the Spatial Compress module is introduced to filter and compress large-scale patch tokens, retaining key information while mitigating the quadratic complexity caused by excessive input tokens. This design greatly improves the model's ability to handle fine-grained tasks while ensuring computational efficiency.

\smallskip
\noindent \textbf{Visual Processing.} Given a video-question pair as input, the video $\bm{V}=[\bm{v}_0, \bm{v}_1, \dots, \bm{v}_{n}]$ comprises $n$ frames. The first step is to extract visual features from each frame using a language-supervised encoder, such as the CLIP visual model. In most implementations, the CLIP encoder is kept frozen during fine-tuning to preserve its original representational capabilities. Similarly, our VideoExpert model utilizes a frozen CLIP ViT-L/14~\cite{CLIP} as the visual encoder. Each frame is processed independently through the visual encoder to extract features:

\begin{equation}
    \left\{\bm{f}_i^{cls}, \bm{f}_i^{1}, \bm{f}_i^{2}, \dots, \bm{f}_i^{p} \right\}= \text{ViT}(\bm{v}_i),
\end{equation}

\noindent where $i=\left\{0, 1, \dots, {n}\right\}$. The class token $\bm{f}_i^{cls}$ encodes the semantic information of the $i$-th frame. Meanwhile, $\bm{f}_i^{p}$ represent features extracted from various local patches within the frame, providing fine-grained details. Here, $p$ denotes the number of patches in each frame.

Most image-based MLLMs use all patch tokens as input, providing comprehensive and detailed visual information. Unfortunately, videos typically have orders of magnitude more tokens than images. As a result, researchers are compelled to downsample video to extremely low frame rates, which significantly degrades model performance on temporally sensitive tasks. To this end, our goal is to preserve both temporal context and spatial detail, within a computationally efficient framework. Specifically, we use low-resolution, high-frame-rate T-tokens to simulate a video's complete temporal context. 

\begin{equation}
    {\bm{z}}_T^{i} = \bm{f}_i^{cls}, i=\left\{0, 1, \dots, {n}\right\}
\end{equation}

\noindent where ${\bm{z}}_T^{i}$ represent T-tokens. Recognizing that videos often contain redundant tokens, the Spatial Compress module is proposed to generate S-tokens from a large number of patch tokens to capture as much valuable spatial detail as possible. 

\vspace{-1 em}
\begin{equation}
    {\bm{z}}_S^{1}, {\bm{z}}_S^{2},\dots, {\bm{z}}_S^{m} = h_{\psi}(\bm{f}_1^{1}, \bm{f}_1^{2}, \dots, \bm{f}_1^{p}, \dots, \bm{f}_n^{p}),
\end{equation}

\noindent $h_{\psi}(\cdot)$ is the Spatial Compression module, which outputs a totally of $m$ S-tokens. $m \ll p \times n$. This design preserves both spatial and temporal information, combining them to deliver a robust video representation for VideoExpert.

\begin{equation}
    {\bm{Z}} = [{\bm{z}}_T^{1}, {\bm{z}}_T^{2},\dots, {\bm{z}}_T^{n}, {\bm{z}}_S^{1}, {\bm{z}}_S^{2},\dots, {\bm{z}}_S^{m}].
\end{equation}

Lastly, the T-token and S-token are mapped to the same embedding space as the LLM through a vision-language adapter. The final visual features are denoted as:

\begin{equation}
    \bm{X}_{visual} = g_{\varphi}({\bm{Z}}).
\end{equation}

\noindent where $\bm{X}_{visual} \in {{\mathbb{R}} ^ {(n+m) \times d}}$ is the visual sequence that LLM can comprehend. The adapter $g_{\varphi}$ is implemented as a linear layer. $d$ denotes the hidden dimension of LLM.

\smallskip
\noindent\textbf{LLM Input.} After visual processing, video features are concatenated with text tokens and jointly used as input to the LLM. The LLM $\mathcal{F}_{LLM}$ generates a sequence response as output. This procedure can be expressed as:

\begin{equation}
    \bm{Y}_{response} = \mathcal{F}_{LLM}(\bm{X}_{visual}, \bm{X}_{question} ).
\end{equation}

\noindent The hybrid input allows various temporal-sensitive video tasks to be reframed as language-based instructions and responses. For example, temporal grounding can be solved by prompting the model with questions like, ``During which frames \{\} happened?". Similarly, dense captioning can be implemented by asking the model to generate multiple sentences along with their corresponding start and end times. Moreover, VideoExpert can handle standard video question answering while providing relevant visual evidence.

\smallskip
\noindent \textbf{LLM Output.} Existing methods typically generate absolute timestamps or relative time tokens, but struggle with performance due to excessive reliance on learned language patterns. The proposed VideoExpert takes a different approach by offloading temporal grounding from content generation, delegating it entirely to the Temporal Expert. To achieve this, the original LLM vocabulary is extended with a new token, \texttt{<LOC>}, which explicitly signals a request for localization output. The textual response $\bm{Y}_{response}$ generated by VideoExpert focuses solely on content. When localization is required, the LLM generates a \texttt{<LOC>} token instead of implanting timestamps within the text. The feature associated with the \texttt{<LOC>} token is extracted and sent to the Temporal Expert as a prompt, guiding it to generate the timestamps. This design eliminates explicit timestamps within the text response, preventing the model from relying on text to predict timestamps. By decoupling temporal localization from textual generation and allowing each component to specialize in its task, this framework enhances the precision of timestamp predictions and improves the quality of textual responses.

\subsection{Temporal \& Spatial Expert}

To address the challenges of managing temporal information in videos, VideoExpert leverages expert collaboration. The Temporal Expert is responsible for temporal perception and grounding, consisting of an adapter within the LLM and a lightweight temporal expert head. The adapter, implemented using LoRA~\cite{lora}, equips the model with temporal perception by processing T-tokens. The temporal expert head works closely with the adapter, utilizing T-tokens encoded jointly by the adapter and the LLM to achieve accurate temporal grounding. Meanwhile, the Spatial Expert focuses on generating content-related responses. It takes S-tokens and text tokens as input, following instructions while providing guidance to the Temporal Expert. This design ensures seamless coordination between temporal localization and content generation.

Specifically, the input visual-text features are defined as $ \bm{X} = [\bm{X}_{visual}, \bm{X}_{question}] \in \mathbb{R}^{(n+m+n_q) \times d} $ , where $\bm{X}_{visual} = [\bm{X}_{T}, \bm{X}_{S}]$ consist of T-token and S-token. $ n_q $ represents the number of text tokens. The adapter of the Temporal Expert accepts only T-tokens as input and leverages residual modeling to fine-tune a small set of parameters, enabling the model to develop temporal perception.

\vspace{-1 em}

\begin{equation}
    [{\bm{X}}_T^{\prime}, {\bm{X}}_S^{\prime}, {\bm{X}}_{question}^{\prime}] = \bm{W}_o [\bm{X}_{T}, \bm{X}_{S}, \bm{X}_{question} ],
\end{equation}

\vspace{-1 em}

\begin{equation}
    {\bm{X}}_T^{\prime} = {\bm{X}}_T^{\prime} + \Delta\bm{W}_T \bm{X}_{T},
\end{equation}

\vspace{-.3em}

\noindent $\bm{W}_o$ is the original parameters of the LLM, which remain frozen at all times. $\Delta\bm{W}_T$ denotes the trainable parameters of the Temporal Expert's adapter. Similarly, the adapter of the Spatial Expert processes text tokens along with S-tokens, empowering the model with instruction-following capabilities and enabling it to generate content-related responses.

\vspace{-.3em}

\begin{equation}
% \begin{aligned}
[{\bm{X}}_S^{\prime}, {\bm{X}}_{question}^{\prime}] = [{\bm{X}}_S^{\prime}, {\bm{X}}_{question}^{\prime}] 
 + \Delta\bm{W}_S [\bm{X}_{S}, \bm{X}_{question}],
% \end{aligned}
\end{equation}

\noindent where $\Delta\bm{W}_S$ corresponds to the trainable parameters of adapter of the Spatial Expert.

The model generates text tokens as responses as usual. However, when the \texttt{<LOC>} token appears in $\bm{Y}_{response}$, it signals a request for localization output. The final-layer embedding of the \texttt{<LOC>} token from the LLM is extracted and processed through a multi-layer perceptron to produce \( \bm{h}_{loc} \). Subsequently, the T-token, jointly encoded by the Temporal Expert and the LLM, are fed into the temporal expert head along with $\bm{h}_{loc}$ to generate the final localization result.

\begin{equation}
    \bm{X}_T^{\prime\prime} = \phi ({\bm{X}}_T^{\prime}, \bm{h}_{loc}).
\end{equation}

\noindent The temporal expert head first reweights the T-tokens using \( \bm{h}_{loc} \), as defined in function (10). It then completes the grounding task through two branches: the indicator branch, which estimates the probability of each T-token being classified as foreground or background, and the boundary branch, which predicts the offsets of each T-token relative to the ground truth. Specifically, the indicator branch comprises three $1\times 3$ convolutional layers, each with $d$ filters and followed by a ReLU activation function. Finally, a sigmoid activation layer is attached to output the predictions $\tilde{k}_i$ per frame. The boundary branch is designed with a similar architecture to the indicator branch, except that the final layer has $2$ output channels for predicting the left and right offsets. Given \( \bm{X}_T^{\prime\prime} \in \mathbb{R}^{n \times d} \) as input, this branch generates per-frame offsets \( \{ \tilde{d}_i \}_i^{n} \).

\subsection{Spatial Compress}

We have explored the use of T-tokens to simulate the full temporal context of video. However, excessively compressed T-tokens lead to a substantial loss of detail. Although dense visual patches can provide richer information, feeding all patches into the MLLM incurs high computational costs. Fortunately, video has a much lower information density than language input. The core goal of the proposed Spatial Compression module is to reduce redundancy while retaining visual detail.

\begin{figure}[!t]
  \centering
  % \fbox{\rule{0pt}{1.5in} \rule{1\linewidth}{0pt}}
   \includegraphics[width=0.9\linewidth]{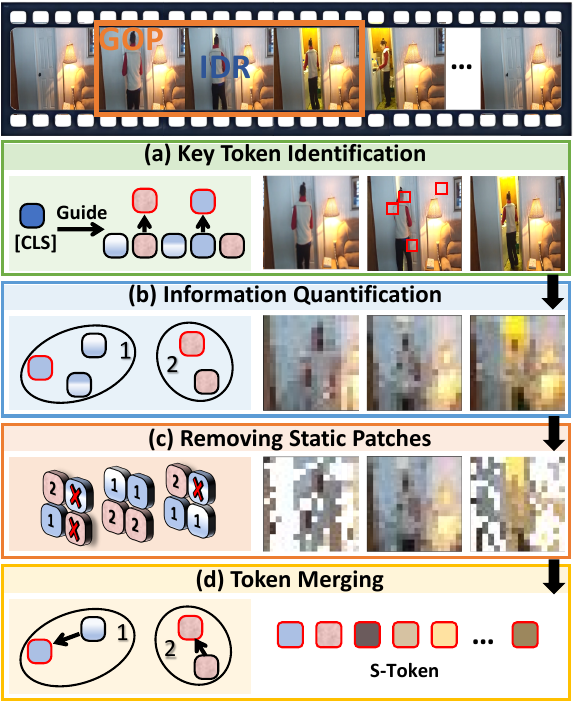}
   \caption{The pipeline of the Spatial Compress module.}
   \label{spatial_compress}
   \vspace{-0.5 cm}
\end{figure}

Inspired by modern video compression techniques such as H.264~\cite{h264}, we compress information with the help of both intra- and inter-frame contexts, as shown in Fig.~\ref{spatial_compress}. Specifically, the video is first divided into Groups of Pictures (GOPs). We uniformly sample $u$ frames from the video as IDR-frames and then expand the GOP boundaries forward and backward based on similarity, resulting in $u$ GOPs. Within each GOP, the IDR-frame is compressed independently, while the remaining frames, treated as P-frames, are compressed using inter-frame methods by referencing the corresponding IDR-frame. The following four steps are performed sequentially to compress tokens: \textbf{(a) Key Token Identification.} We select certain tokens within the IDR-frames as key tokens. First, the CLS token aggregates key information from the entire image, and the patch tokens that receive greater attention from the CLS token are always informative. Therefore, we utilize the attention scores of the CLS token to select these tokens as key tokens. Additionally, to preserve contextual information, we uniformly sample a few tokens from the remaining as context tokens, ensuring that no potentially important details are omitted. \textbf{(b) Information Quantification.} We identify each token within the GOP that conveys information similar to the selected key tokens and group them accordingly. \textbf{(c) Removing Static Patches.} We detect static information that repeats over time in this step, enabling the model to reduce redundancy and focus more on dynamic content. Specifically, we define two consecutive patches located at the same spatial coordinates $(x, y)$ and different time positions $t_1$ and $t_2$ (where $t_2 = t_1 + \Delta t$). If their class labels are identical, these tokens are considered temporally repetitive. Such tokens are removed from the P-frames and retained only in the IDR-frame. \textbf{(d) Token Merging.} Finally, we merge tokens that belong to the same category across frames, creating S-tokens to be further used in the spatial expert.

\subsection{Boundary-aware Training Paradigm} 

Besides the architecture, training tasks and data are crucial in shaping the MLLM. Our training data comprises three components, all sourced from public datasets. 

\begin{comment}

\begin{figure}[!tp]
  \centering
  \includegraphics[width=\linewidth]{Photo/ShowCase_3.pdf}
  \caption{{An example illustrating the temporal-sensitive video understanding tasks addressed by VideoExpert, which include Video Temporal Grounding (Green), Video Highlight Detection (Blue), Grounding Video Question Answering (Orange), and Dense Video Captioning (Yellow).}}
  \label{fig_showcase}
  \vspace{-0.5 cm}
\end{figure}
\end{comment}

\smallskip
\noindent \textbf{• Temporal Grounding.} The goal of this task is to localize the event described by a question sentence within a video. To generate data matching the format of visual question answering, one question-answer template is employ like,

% Temporal grounding datasets typically consist of videos and their corresponding questions. 

\vspace{-0.3 em}

 \begin{table}[h!]\centering
\begin{minipage}{0.99\columnwidth}
% \vspace{-2em}    
\centering
\begin{tcolorbox} 
    \raggedright
    \small
     % \hspace{-6mm}
    % $\Xmat_{\texttt{system}}$  \PredSty{\texttt{<EOS>}}\\
    USER: \texttt{<VIDEO>} During which frames \texttt{<Question>} happend? 
    {VideoExpert}: During \texttt{<LOC>}.
    % \PredSty{\texttt{<EOS>}} $\textbackslash \texttt{n}$  \\
    % \texttt{<EOS>} \\
\end{tcolorbox}
\label{tab:input_sequence}
\end{minipage}
% \vspace{-2em} 
\end{table}

% \vspace{-0.5 em}

\noindent Here, \texttt{<Question>} represents the question text, while \texttt{<VIDEO>} serves as a placeholder for the tokens of visual input. The \texttt{<LOC>} is replaced with actual timestamps to serve as the final result.

\smallskip
\noindent \textbf{• Dense Video Captioning.} Each video is described by a set of sentences, with each sentence accompanied by its corresponding start and end timestamps of the event. This structure allows the data to be easily converted into question-answer pairs using a template like:

\begin{table}[h!]\centering
\begin{minipage}{0.99\columnwidth}
% \vspace{-2em}    
\centering
\begin{tcolorbox} 
    \raggedright
    \small
     % \hspace{-6mm}
    % $\Xmat_{\texttt{system}}$  \PredSty{\texttt{<EOS>}}\\
    USER: \texttt{<VIDEO>} Describe the provided video in detail. Each sentence should begin with the timestamps. 
    {VideoExpert}: During \texttt{<LOC>}, xxx. During \texttt{<LOC>}, xxx.
    % \PredSty{\texttt{<EOS>}} $\textbackslash \texttt{n}$  \\
    % \texttt{<EOS>} \\
\end{tcolorbox}
\label{tab_DVC}
\end{minipage}
\end{table}

\vspace{-1em}
\noindent The language loss is used to constrain the model to generate outputs that adhere to a predefined template and align with the intended content. Each timestamp serves as the ground truth to supervise the corresponding grounding results. During training, additional templates are also utilized to generate QA data, ensuring diversity in the training dataset.

\smallskip
\noindent \textbf{• Video Question Answering.} To maintain the original VQA capabilities of the MLLM, we incorporate the VQA dataset during training. Following LLaVA~\cite{llava}, the question-answering task is already represented as language instructions. These datasets do not include any grounding question-answering samples that require temporal reasoning. Surprisingly, even without training on complex reasoning data, VideoExpert exhibits impressive zero-shot performance on reasoning-based temporal grounding tasks, such as Grounding QA.

\begin{comment}

\begin{table}[h!]\centering
\begin{minipage}{0.99\columnwidth}
% \vspace{-2em}    
\centering
\begin{tcolorbox} 
    \raggedright
    \small
     % \hspace{-6mm}
    % $\Xmat_{\texttt{system}}$  \PredSty{\texttt{<EOS>}}\\
    USER: \texttt{<VIDEO>} During which frames \texttt{<Question>} happend? 
    {VideoExpert}: During \texttt{<LOC>}.
    % \PredSty{\texttt{<EOS>}} $\textbackslash \texttt{n}$  \\
    % \texttt{<EOS>} \\
\end{tcolorbox}
\label{tab:input_sequence}
\end{minipage}
\end{table}

\vspace{-1em}

\end{comment}

\subsection{Training Objectives.}

The model is trained end-to-end using the text generation loss $\mathcal{L}_\mathrm{text}$ and the boundary loss $\mathcal{L}_\mathrm{ce} + \mathcal{L}_\mathrm{b}$. The general objective $\mathcal{L}$ is defined as:

\begin{equation}
\mathcal{L} = \lambda_\mathrm{text} \mathcal{L}_\mathrm{text} + \mathcal{L}_\mathrm{ce} + \mathcal{L}_\mathrm{b} ,
\end{equation}

\noindent where $\lambda_\mathrm{text} \in {\mathbb{R}}$ are the trade-off hyperparameters. $\mathcal{L}_\mathrm{text}$ is the auto-regressive cross-entropy loss used for text generation. The model is driven by the language generated by the LLM. We adopt the standard language modeling training objective, which is defined as follows:

\begin{equation}
% \begin{split}
    \mathcal{L}_\mathrm{text} = \sum_{i=1}^{L}\log p( {\bm{y}_{i}} | \bm{X}, \bm{Y}_{response, <i}),
% \end{split}
\label{eq:auto_regressive}
\end{equation}

\noindent where $L$ the length of the response sequence of the model. The boundary loss encourages the model to generate high-quality localization results. Specifically, $\mathcal{L}_\text{ce}$ represents the cross-entropy loss used to evaluate whether the predicted $\tilde{k}_i$ is correctly classified as foreground or background.

\begin{equation}
    \mathcal{L}_\text{b} =
    \lambda_\text{L1}\mathcal{L}_{\text{L1}}\left(\tilde{d}_i, {d_i}\right)+
    \lambda_\text{iou} \mathcal{L}_\text{iou}\left( \tilde{b}_i, {b_i} \right).
\end{equation}

\noindent $\mathcal{L}_\text{b}$ quantifies the discrepancy between the Ground-Truth moment and the predicted moment. We derive the boundary $\tilde{b_i}$ from the predicted offset and employ a combination of smooth $L1$ loss and generalized IoU loss as the training objectives.

% === IV. Transistor Class-F inv Rectifier ========================================
% =================================================================================
% \newpage
%%%%%%%%%%%%%%%%%%%%%%
% Experiments
%%%%%%%%%%%%%%%%%%%%%
% \newpage

\section{Experiments}

In this section, we structure our experiments to investigate the following questions: 
\begin{itemize}
    \item \textbf{Q1.} Is it possible to get a general model that is good at multiple temporal-sensitive video tasks?
    \item \textbf{Q2.} Does VideoExpert achieve better performance compared to current MLLMs across various setups?% Will VideoExpert achieve better zero-shot performance compared to current MLLMs across various setups?
    \item \textbf{Q3.} Are the proposed modules effective? We perform ablation studies to examine different configurations in VideoExpert, such as Temporal Expert. % and Spatial Compression.
    \item \textbf{Q4.} Can a model that excels in temporal-sensitive tasks also perform well in text generation?
\end{itemize}

\subsection{Datasets and Settings}

We pre-trained VideoExpert on multiple tasks, as summarized in Tab~\ref{datasets}. All datasets used are sourced from widely used public repositories. Specifically, \textbf{InterVid}~\cite{InternVid} is a video-centric dataset designed to support multimodal understanding and generation. It employs a multi-scale approach to autonomously generate high-quality video-text descriptions. Following~\cite{VTimeLLM}, a subset of around 100K videos with temporal annotations and event descriptions was selected. This dataset is primarily used for training in temporal grounding and dense captioning. \textbf{ActivityNet-Captions}~\cite{dvc_twostage_2} consists of 20K videos of human activities, with each video averaging 120 seconds and annotated with 3.65 temporally localized sentences. The description uniquely corresponds to a single segment. In this paper, only the training set is employed for training temporal grounding and dense captioning tasks. \textbf{Next-QA}~\cite{nextQA}is a video question-answering benchmark containing 5.4K videos and 47K question-answer pairs. The questions focus on causal action reasoning, temporal action analysis, and general scene understanding. We primarily adopt multiple-choice QA tasks to populate question-answer templates.

VideoExpert is evaluated on four datasets spanning various tasks. For each setting, we briefly introduce the dataset and the corresponding evaluation metrics.

\begin{itemize}

\item \textbf{Charades-STA}~\cite{twostage1} consists of $6,672$ videos and provides $16,124$ query-moment pairs for the video temporal grounding task. The average length of the videos and target moments are $30.60$ and $8.09$ seconds, respectively. Following previous work, $12,404$ query-moment pairs are utilized for training and $3,720$ for testing. 

\vspace{0.2em}
\textbf{Metric:} Evaluation metrics in this benchmark include Recall@1 with IoU thresholds (R1\text{@}\{0.3, 0.5, 0.7\}) and mean IoU (mIoU). Higher IoU values indicate more precise moment matching.

\vspace{0.2em}
\item \textbf{QVhighlights}~\cite{moment-detr} is a recent benchmark designed for video temporal grounding and highlight detection based on natural language queries. It consists of $10,310$ samples annotated with human-written text queries. Each query is associated with multiple moments within a video. Experiments are conducted on the standard split, i.e., $7,218$ query-moment pairs for training, $1,550$ for validation, and $1,512$ for testing. Notably, QVHighlights provides a fair evaluation, as the evaluation of the test split results requires submission to the server. 

\vspace{0.2em}
\textbf{Metric:} Following the convention, $\text{R1@}\{0.5, 0.7\}$, $\text{mAP@}\{0.5, 0.75\} $, and average mAP (mAP\text{@}Avg) are used for evaluating temporal grounding. For highlight detection, the metrics of mAP and Hit@1 are employed, with the thresholds set to ``Very Good''.

\begin{table}[!t]
\centering
\footnotesize
\caption{{\upshape\small{ \textbf{Dataset statistics.} The datasets listed on the upper section are used for pre-training, while the datasets on the lower section are used for downstream tasks. TG: Temporal Grounding, HD: Highlight Detection, DVC: Dense Video Captioning, GQA: Grounding Question Answer.}}}
% \vspace{-0.5em}
\renewcommand{\arraystretch}{1.1}
\setlength{\tabcolsep}{3pt}
\begin{tabular}{lcccc}
\toprule
\multicolumn{1}{l}{\textbf{Datasets}} & \textbf{\# Samples} & \textbf{Task} & \textbf{Video Len.} & \textbf{Video Domain}                          \\ \midrule
\multicolumn{3}{l}{\textit{\textbf{Pre-training}}}                               \\ [1.5pt]
InterVid                     & 107.3K     & TG\&DVC    & 60s   & Web                                  \\
ActivityNet                  & 28.2K      & TG\&DVC   & 120s    & Daily                                  \\
% DiDeMo                       & 10K      & TG\&DVC   & s    & Daily                                  \\
Next-QA                      & 35.1K      & QA   & 42s    & Daily                                  \\\midrule
\multicolumn{3}{l}{\textit{\textbf{Fine-tuning and/or Eval.}}}                        \\ [1.5pt]
Charades-STA                 & 16.1K         & TG    &30s & Indoor                           \\
QVHighlights                 & 10.3K    & \multicolumn{1}{c}{TG\&HD} &150s & VLog, News      \\
YouCookII                    & 1.7K        & DVC    &320s & Cooking                          \\
Next-GQA                      & 43.1K    & \multicolumn{1}{c}{GQA}    &42s & Daily    \\ 
\bottomrule
\end{tabular}
\label{datasets}
\vspace{-0.5 em}
\end{table}

\begin{figure*}[t]
 \begin{minipage}[t]{0.422\textwidth}
  \centering
  \scriptsize
\renewcommand{\arraystretch}{1.1}
\renewcommand\tabcolsep{0pt}
\footnotesize
% \caption{{\upshape \textbf{Temporal Grounding results on Charades-STA test split.} PT: pre-training; ZS: zero-shot inference.}}
\makeatletter\def\@captype{table}\makeatother\caption{\small{\textbf{Temporal Grounding results on Charades-STA test split.} PT: pre-training; ZS: zero-shot inference.}}
% \vspace{-0.5em}

\begin{tabularx}{\linewidth}{@{}p{2.15cm}p{0.75cm}<{\centering}p{0.6cm}<{\centering}p{0.6cm}<{\centering}p{0.88cm}<{\centering}p{0.88cm}<{\centering}p{0.88cm}<{\centering}p{0.9cm}<{\centering}}
\toprule
\multicolumn{1}{c}{\multirow{3}{*}{\textbf{Method}}} & \multirow{3}{*}{\textbf{Style}} & \multirow{3}{*}{\textbf{PT}} & \multirow{3}{*}{\textbf{ZS}} & \multicolumn{4}{c}{\textbf{Temporal Grounding}}  \\ \cline{5-8}
\multicolumn{1}{c}{}                        &                        &                     &                     & \multicolumn{3}{c}{R1} & \multirow{2}{*}{mIoU} \\ \cline{5-7}
\multicolumn{1}{c}{}                        &                        &                     &                     & @0.3   & @0.5  & @0.7  &                       \\ \midrule
2D-TAN \cite{2D-TAN}                                      & Spec.                  & -                   & -                   & 58.8  & 46.0 & 27.5 & 41.3                 \\
VSL-Net \cite{vslnet}                                     & Spec.                 & -                   & -                   & 60.3  & 42.7 & 24.1 & 41.6                 \\
M-DETR \cite{moment-detr}                                      & Spec.                  & -                   & -                   & 65.8  & 52.1 & 30.6 & 45.5                 \\ \midrule \midrule
Momentor \cite{Momentor}                                    & Gen.                   & \checkmark          & \checkmark          & 42.6   & 26.6   & 11.6   & 28.5                   \\
TimeChat \cite{timechat}                                    & Gen.                   & \checkmark          & \checkmark          &  46.0     & 29.4  & 12.9  & 30.4                      \\
VTimeLLM \cite{VTimeLLM}                                    & Gen.                   & \checkmark          & \checkmark          & 51.0   & 27.5  & 11.4  & 31.2                 \\
VTGLLM \cite{vtgllm}                                      & Gen.                   & \checkmark          & \checkmark          & 52.0   & 33.8   & 15.7  & -                      \\
HawkEye \cite{hawkeye}                                     & Gen.                   & \checkmark          & \checkmark          & 50.6   & 31.4   & 14.5  & 33.7                      \\ \midrule

\textbf{Ours}                               & Gen.                   & \checkmark                 & \checkmark          & {61.5}  & {40.3} & {20.9} & {41.1}                 \\
\textbf{Ours}                               & Gen.                   & \checkmark                 & -                   & \textbf{74.3}  & \textbf{60.8} & \textbf{36.5} & \textbf{52.2}                 \\ \bottomrule
\end{tabularx}
% \vspace{-1em}
% \captionsetup{font={small}}
\label{MR_Charades}
  \end{minipage}
  \hspace{0.25cm}
  \begin{minipage}[t]{0.565\textwidth}
   \centering
   \scriptsize
\renewcommand{\arraystretch}{1.1}
\renewcommand\tabcolsep{0pt}
\footnotesize
% \caption{{\upshape \textbf{Jointly Temporal Grounding and Highlight Detection results on QVHighlights test split.} PT: pre-training; ZS: zero-shot inference.}}
\makeatletter\def\@captype{table}\makeatother\caption{\small{\textbf{Jointly Temporal Grounding and Highlight Detection results on QVHighlights test split.} PT: pre-training; ZS: zero-shot inference.}}
% \vspace{-0.5em}

\begin{tabularx}{\linewidth}{@{\hspace{0mm}}p{2cm}p{0.75cm}<{\centering}p{0.62cm}<{\centering}p{0.62cm}<{\centering}p{0.9cm}<{\centering}p{0.9cm}<{\centering}p{0.9cm}<{\centering}p{0.9cm}<{\centering}p{0.9cm}<{\centering}p{0.9cm}<{\centering}p{0.9cm}<{\centering}}
\toprule
\multicolumn{1}{c}{\multirow{3}{*}{\textbf{Method}}} & \multirow{3}{*}{\textbf{Style}} & \multirow{3}{*}{\textbf{PT}} & \multirow{3}{*}{\textbf{ZS}} & \multicolumn{5}{c}{\textbf{Temporal Grounding}}                           & \multicolumn{2}{c}{\textbf{HD}}               \\ 
\cline{5-11}
\multicolumn{1}{c}{}                        &                        &                     &                     & \multicolumn{2}{c}{R1} & \multicolumn{3}{c}{mAP} & \multicolumn{2}{l}{$\ge $ Very Good} \\ \cline{5-11} 
\multicolumn{1}{c}{}                        &                        &                     &                     & @0.5       & @0.7      & @0.5   & @0.75  &  Avg. & mAP               & HIT1            \\ \midrule
M-DETR \cite{moment-detr}                                      & Spec.                  &\checkmark           &\checkmark                     & 2.45       & 0.58      & 1.6    & 0.33   & 0.52  & 26.12             & 31.61            \\
M-DETR \cite{moment-detr}                                      & Spec.                  &\checkmark           & -                   & 59.78      & 40.33     & 60.51  & 35.36  & 36.14 & \textbf{37.43}             & 60.17            \\
M-DETR \cite{moment-detr}                                      & Spec.                  & -                   & -                   & 52.89      & 33.02     & 54.82  & 29.40  & 30.73 & 35.69             & 55.60            \\ \midrule \midrule
Momentor \cite{Momentor}                         & Gen.                   & \checkmark           & \checkmark           & 17.00             & -           & -        & -        & -        & 7.60                   & -                  \\
TimeChat \cite{timechat}                                    & Gen.                   &\checkmark           &\checkmark           &9.92            &4.86           &7.49        &3.03        &3.72       &14.37                   &23.92                  \\
VTimeLLM \cite{VTimeLLM}                                    & Gen.                   &\checkmark           &\checkmark           & 49.81      & 30.32     & 40.58  & 22.73  & 22.86 & -                 & -                \\
VTGLLM \cite{vtgllm}                                        & Gen.                   &\checkmark           &\checkmark           & -            & -           & -        & -        & -       &16.50             & 33.50                 \\
SeViLA \cite{sevila-twostage}                         & Gen.                   & -           & -           & 54.50            &36.50           & -        & -        & 32.30       & -                   & -                  \\ \midrule
\textbf{Ours}                                      & Gen.                   &\checkmark           &\checkmark           & 54.77      & 35.35     & 53.61  & 30.97  & 31.06 & 35.76             & 52.71            \\
\textbf{Ours}                                      & Gen.                   & \checkmark          & -                   & \textbf{67.23}      & \textbf{47.81}     & \textbf{63.09}  & \textbf{40.50}  & \textbf{39.62} & {36.13}             & \textbf{60.97}           \\\bottomrule
\end{tabularx}

% \vspace{-1em}
% \captionsetup{font={small}}
% \makeatletter\def\@captype{table}\makeatother\caption{\small{\textbf{\HL~results of Top-5 mAP on TVSum.} $\dagger$ denotes using audio modality.}}
\label{MR_QVHL}
\end{minipage}
\vspace{-0.5 em}
\end{figure*}

% \textsuperscript{\boldsymbol{$\uparrow$}}

\vspace{0.2em}
\item \textbf{YouCookII}~\cite{YouCookII} is a dataset designed for video description generation, containing over 2,000 untrimmed cooking videos from YouTube, totaling more than 176 hours. On average, each video lasts 320 seconds and is annotated with 7.7 temporally-localized sentences. The videos are segmented into multiple clips, each associated with specific timestamps and detailed descriptions. 

\vspace{0.2em}
\textbf{Metric:} SODA\textsubscript{c} is a metric designed for this task, which evaluates the captions generated while considering the storyline of the video. Temporal alignment between the generated events and the Ground-Truth is also taken into account. Captioning metrics, such as CIDE\textsubscript{r} and METEOR, are calculated based on these matched pairs.

\vspace{0.2em}
\item \textbf{Next-GQA}~\cite{NextGQA} is a new benchmark for Grounding QA, created by adding temporal annotations to Next-QA. It challenges VLMs to answer questions while providing visual evidence. This setup aims to determine whether model predictions are based on relevant video content or influenced by spurious correlations in language and irrelevant visual context. 

\vspace{0.2em}
\textbf{Metric:} Evaluation metrics for this benchmark include two parts. For visual evidence grounding, metric such as IoP and  IoU assess whether the predicted temporal window aligns with the ground truth. For question answering, results are reported as the percentage of correctly answered questions(Acc@QA). Additionally, a grounded QA accuracy metric (Acc@GQA) is defined, measuring the percentage of questions that are both correctly answered and visually grounded (IoP $\geq$ 0.5).
     
\end{itemize}

\noindent
\textbf{Implementation Details.} In our study, Vicuna-1.5 7B~\cite{vicuna} is used as the Large Language Model. For each dataset, we sample 100 frames per video. A total batch size of 128 is used throughout the training process.  The AdamW optimizer is employed with a cosine learning rate decay and a warm-up period. Both expert LoRA~\cite{lora} configurations use a rank of 32 and an alpha of 64. The maximum response length for the model is set to 512 by default.

\subsection{Main Results}
In this section, the VideoExpert is compared with the state-of-the-art methods on four temporal-sensitive benchmarks.

\smallskip
\noindent\textbf{Temporal Grounding and Highlight Detection.} The evaluation begins with two common temporal-sensitive video tasks. Table~\ref{MR_Charades} presents the performance of VideoExpert on the Charades-STA benchmark for the temporal grounding task. Table~\ref{MR_QVHL} compares its performance on the QVHighlights test split, covering joint temporal grounding and highlight detection tasks. Notably, the QVHighlights benchmark offers an official online test evaluation, ensuring reliable result reporting. 

Overall, our method achieves superior zero-shot performance, surpassing all previous LLM-based approaches and even rivaling specialized methods. After fine-tuning, VideoExpert outperforms several classic specialized models across two tasks. Specifically, our method attains an mIoU accuracy of 41.1 on the Charades-STA benchmark, significantly surpassing the baseline VTimeLLM by 10.1 points. Compared to the recently proposed VTGLLM, our method demonstrates a 9.5 points improvement in R1@0.3 and a 5.2 points gain in the more stringent R1@0.7 metric. For the QVHighlights benchmark, VideoExpert consistently have improvements across tasks. In particular, it outperforms VTGLLM by over 19 points in both mAP and HIT@1 for the highlight detection task. Furthermore, after fine-tuning on the benchmark’s training set, VideoExpert achieves an mIoU of 52.2 on Charades, surpassing most classic supervised specialized models. These results highlight significant and comprehensive improvements across both benchmarks, underscoring the effectiveness of our proposed approach.

\begin{table}[!t]
\footnotesize
\centering
\caption{{\upshape \small{\textbf{Dense Video Captioning results on YouCookII.} PT: pre-training; ZS: zero-shot inference.}}}
% \vspace{-0.5em}
\renewcommand{\arraystretch}{1.1}
\setlength{\tabcolsep}{4 pt}
\begin{tabular}{lcccccc}
\toprule
\multicolumn{1}{c}{\multirow{1}{*}{\vspace{0cm}\textbf{Method}}}     & \multirow{1}{*}{\vspace{0cm}\textbf{Style}} & \multirow{1}{*}{\vspace{0cm}\textbf{PT}} & \multirow{1}{*}{\vspace{0cm}\textbf{ZS}} & \multicolumn{1}{c}{\textbf{SODA\textsubscript{c}}$\uparrow$} & \multicolumn{1}{c}{\textbf{CIDE\textsubscript{r}}$\uparrow$} & \multicolumn{1}{c}{\textbf{METEOR}$\uparrow$}                                           \\ \cmidrule{1-7}
% \multicolumn{1}{c}{}                            &                 &       & &  \\ \midrule
% Vid2Seq*~\cite{}                                        & Spec.        &  \checkmark       & 2.4                         & 10.1                      & 3.3                        \\ 
MT~\cite{MT}                                              & Spec.        &  -             & -   & -                           & 6.1                       & 3.2                        \\
Vid2Seq~\cite{vid2seq}                                         & Spec.        &  -             & -   & 4.0                         & 18.0                      & 4.6                        \\ \midrule \midrule
% Vid2Seq~\cite{}                                         & Spec.        &  \checkmark    & -   & 7.9                         & 47.1                      & 9.3                        \\ \midrule \midrule

VideoChat~\cite{videochat2}                                      & Gen.         &  \checkmark   & \checkmark    & 0.2                         & 0.6                       & -                          \\
TimeChat~\cite{timechat}                                        & Gen.         &  \checkmark   & \checkmark    & 1.2                         & 3.4                       & -                          \\
VTGLLM~\cite{vtgllm}                                         & Gen.         &  \checkmark   & \checkmark    & 1.5                         & 5.0                      & 1.9                           \\
VTimeLLM~\cite{VTimeLLM}                                        & Gen.         &  \checkmark   & \checkmark    & 0.9                         & 3.4                       & 1.1                       \\
\textbf{Ours}                                          & Gen.         &  \checkmark    & \checkmark    & \textbf{2.1}                    & \textbf{6.0}               & \textbf{2.7}                        \\ \midrule

TimeChat~\cite{timechat}                                        & Gen.         &  \checkmark    & -    & 3.4                         & 11.0                      & -                          \\
VTGLLM~\cite{vtgllm}                                         & Gen.         &  \checkmark   & -     & 3.6                         & 13.4                      & -                           \\
\textbf{Our}                                            & Gen.         &  \checkmark   & -     & \textbf{4.2}                             & \textbf{18.7}                          & \textbf{4.8}                            \\
% \textbf{Ours w/Knowl.}                          & Gen.         &  \checkmark       & 4.4        & 18.8      & 4.9       \\  

\bottomrule

\end{tabular}
\label{DVC_Youcook}
\vspace{-1 em}
\end{table}

% \textsuperscript{{\boldsymbol{$\uparrow$}}}

% \multicolumn{5}{l}{\textit{\textbf{Full-training}}} 

% \multicolumn{5}{l}{\textit{\textbf{Zero-Shot}}}       \\ [1.5pt]

% \vspace{-0.2cm}

\begin{table*}[ht]

\footnotesize
\centering
\caption{{\upshape \small{\textbf{Video Grounding Question Answer Results on NExT-GQA.} The pure QA taks result are reported as Acc@QA, representing the percentage of correctly answered questions. Acc@GQA reflects the percentage of questions that are both correctly answered and visually grounded with an IoP $\ge $ 0.5. IoP and IoU are used to evaluate whether the predicted temporal window aligns with the ground truth.}}}
% \vspace{-0.5em}
\renewcommand{\arraystretch}{1.1}
\setlength{\tabcolsep}{5.2 pt}

\begin{tabular}{lcc|cc|cccccc}
\toprule
\textbf{Method}         & \textbf{Style} &\textbf{PT} & Acc@QA$\uparrow$ & Acc@GQA$\uparrow$ & IoP@0.3$\uparrow$ & IoP@0.5$\uparrow$ & mIoP$\uparrow$    & IoU@0.3$\uparrow$ & IoU@0.5$\uparrow$ & mIoU$\uparrow$  \\ \midrule
Random         & Toy   & - & 20.0   & 1.7     & 20.6    & 8.7     & 21.1    & 20.6    & 8.7     & 21.1    \\ 
Human          & Toy   & - & 93.3   & 82.1    & 91.7    & 86.2    & 72.1    & 86.9    & 70.3    & 61.2    \\ \midrule
IGV \cite{igv}            & Spec.       & - & 50.1   & 10.2    & 26.9    & 18.9    & 21.4    & 19.8    & 9.6     & 14.0    \\
Violet-V2 \cite{violetv2}      & Spec.      & \checkmark & 52.9   & 12.8    & 25.1    & 23.3    & 23.6    & 4.3     & 1.3     & 3.1     \\
Temp{[}CLIP{]} \cite{CLIP} & Spec.      & \checkmark & 60.2   & 16.0    & 31.4    & 25.5    & 25.7    & 17.5    & 8.9     & 12.1    \\ \midrule \midrule
FrozenBiLM \cite{frozenbilm}     & Gen.    & \checkmark & 70.8   & 17.5    & 28.5    & 23.7    & 24.2    & 13.5    & 6.1     & 9.6     \\
% Mistral\cite{}        &       & &        &         &         &         &         &         &         &         \\
LLoVi \cite{LLovi-modelchange}         & Gen.    & \checkmark & -       & 11.2    & -    & 20.5    & 20.7    & -    & 6.0    & 8.7    \\
LangRepo \cite{LangRePo-modelchange}         & Gen.    & \checkmark & -   & 11.2    & -    & 20.0    & 20.3    & -    & 6.0    & 8.7    \\
SeViLA \cite{sevila-twostage}         & Gen.    & \checkmark & 68.1   & 16.6    & 34.7    & 22.9    & 29.5    & 29.2    & 13.8    & 21.7    \\
VTimeLLM \cite{VTimeLLM}       & Gen.    & \checkmark & -      & 12.7    & 30.3    & 23.8    & 27.9    & 27.7    & 14.1    & 18.3    \\
VideoStream \cite{videostream}    & Gen.    & \checkmark & -      & 17.8    & -       & \textbf{31.0}     & {32.2}    & -       & 13.3    & 17.8        \\
HawkEye \cite{hawkeye}        & Gen.    & \checkmark & -      & -       & -       &-        & -       & 37.0      & 19.5    & 25.7        \\
\textbf{Ours}                  & Gen.    & \checkmark & \textbf{71.1}    & \textbf{21.6}    & \textbf{45.3}    & 29.3    & \textbf{34.6}    & \textbf{41.0}        & \textbf{22.4}        & \textbf{27.9}        \\ 
\bottomrule
\end{tabular}
\centering
\vspace{-0.5 em}
\label{GQA_nextqa}
\end{table*}

\smallskip
\noindent\textbf{Dense Video Captioning.} This task in YouCook2 presents a significant challenge to a model's multi-task capabilities. The model must accurately localize all events within a given video and generate descriptions that align with the visual content of each event. This requires the model to have strong temporal awareness and content understanding, imposing rigorous demands on its capabilities. The result are shown in Tabel~\ref{DVC_Youcook}. First, VideoChat extract only eight frames as input, making it difficult to achieve precise moment localization. Such imprecision significantly impacts captioning evaluation, with both SODA\textsubscript{c} and CIDE\textsubscript{r} metrics dropping close to zero. In contrast, VideoExpert leverages low-resolution, high-frame-rate T-tokens to simulate the complete temporal context of a video, leading to remarkable performance gains. Furthermore, compared to other LLM-based methods, our model offers three key advantages: (1) More accurate event boundaries and descriptions. Thanks to its split expert architecture, VideoExpert excels in both moment localization and content generation. (2) Comprehensive event capture. VideoExpert effectively identifies all key events in a video while maintaining high descriptive accuracy, as reflected in its high SODA\textsubscript{c} score. (3) Further performance improvements through fine-tuning. After fine-tuning, VideoExpert achieves even greater performance, rivaling most classic specialized methods. Overall, these findings highlight the effectiveness of our approach in Dense Video Captioning task, demonstrating that VideoExpert excels precise temporal grounding and accurate content captioning. This further validating the effectiveness and generalizability of DiffusionVMR.

\begin{table}[!t]
\footnotesize
\centering
\caption{{\upshape \small{\textbf{Effectiveness of different interaction methods between \texttt{<LOC>} and T-tokens on the QVHighlights for temporal grounding(TG) and the Next-GQA for GroundingQA (GQA).}}}}
% \vspace{-0.5em}
\renewcommand{\arraystretch}{1.1}
\setlength{\tabcolsep}{3 pt}

\begin{tabular}{lcccccc}
\toprule
\multicolumn{1}{c}{\multirow{2}{*}{\vspace{-0.2cm}\textbf{Method}}} & \multicolumn{3}{c}{\textbf{TG}} & \multicolumn{2}{c}{\textbf{GQA}} \\ \cmidrule{2-7} 
\multicolumn{1}{c}{}                                 & R1@0.5   & R1@0.7   & mAP@Avg.  & IoU@0.3   & IoU@0.5   & mIoU \\ \midrule
w/o \texttt{<LOC>}                                   & 21.4     & 12.7     & 17.1      & 14.3           & 7.1         & 10.7  \\ \midrule
Add                                                  & \textbf{52.4}     & \textbf{34.9}     & \textbf{31.1}      & \textbf{41.0}  & {22.4}  & \textbf{27.9}\\
Concat.                                              & 50.0     & 34.0     & 30.4      & 40.6       & 21.2    & 26.6      \\
Self-Atten.                                          & 49.7     & 34.7     & 31.0      & 40.7       & \textbf{23.1}    & 27.3      \\ 
\bottomrule
\end{tabular}

\label{ablation_loc}
\vspace{-1 em}
\end{table}

% \textsuperscript{{\boldsymbol{$\uparrow$}}}

% \multicolumn{5}{l}{\textit{\textbf{Full-training}}} 

% \multicolumn{5}{l}{\textit{\textbf{Zero-Shot}}}       \\ [1.5pt]

% \vspace{-0.73cm}\hspace{0.8cm}

\smallskip
\noindent\textbf{Grounding Question Answering.} Beyond simply locating segments based on content descriptions, reasoning-based grounding in response to a given question is an even more challenging yet crucial task. It is a key step toward achieving episodic memory interaction and explainable question answering. To evaluate this capability,  VideoExpert is assessed on the NExT-GQA benchmark.  This task requires the model not only to provide accurate answers based on a given question but also to grounding the relevant video clips that support those answers. The result are presented in Table~\ref{GQA_nextqa}. First, while LLM-based approaches generally excel at question answering, their performance in reasoning grounding remains inconsistent. For example, FrozenBiLM achieves a high GQA accuracy, primarily due to its strong QA capabilities rather than its proficiency in reasoning grounding. Second, VideoExpert achieves the highest IoP and IoU among all compared methods, even outperforming SeViLA, which includes a specialized grounding module. This demonstrates VideoExpert's superior reasoning grounding ability. Finally, the highest Acc@GQA score of VideoExpert further confirms its comprehensive capability in both fine-grained temporal grounding and high-level QA. Unfortunately, despite VideoExpert’s significant advancements in grounding QA, not all correct answers are supported by the appropriate visual evidence. Additionally, we observed that even when the correct visual evidence is identified, it can sometimes lead to wrong responses. There is still room for improvement in this aspect, which future iterations of the model aim to address more effectively.

\subsection{Ablation Studies}
In this section, a series of ablation studies are designed to verify the effectiveness of each component of the proposed approach. All experimental results presented here were obtained without fine-tuning.

\begin{table*}[!t]
\footnotesize
\centering
\caption{{\upshape \small{\textbf{\upshape Effectiveness of the different model components on QVHighlights (TG), YouCookII (DVC) and Next-GQA(GQA).}}}}
% \vspace{-0.5em}
\renewcommand{\arraystretch}{1.1}
\setlength{\tabcolsep}{4 pt}

\begin{tabular}{clccccccccc}
\toprule
\multirow{2}{*}{Row} & \multicolumn{1}{c}{\multirow{2}{*}{Method}} & \multicolumn{3}{c}{\textbf{TG}} & \multicolumn{3}{c}{\textbf{DVC}} & \multicolumn{3}{c}{\textbf{GQA}} \\ \cmidrule{3-11} 
                     & \multicolumn{1}{c}{}                        & R1@0.5$\uparrow$   & R1@0.7$\uparrow$   & mAP@Avg.$\uparrow$ & SODA\textsubscript{c}$\uparrow$  & CIDE\textsubscript{r}$\uparrow$            & METEOR$\uparrow$          & IoU@0.3$\uparrow$     & mIoU$\uparrow$     & Acc@GQA$\uparrow$    \\ \midrule
1                    & Ours                                        &52.4          &\textbf{34.9}          &\textbf{31.1}           & \textbf{2.1}                                  & \textbf{6.0}                & \textbf{2.7}         & \textbf{41.0}         & \textbf{27.9}    & \textbf{21.6}         \\ % \midrule
2                    & -w/o Spatial Compress                       & 52.1      & 34.2     & 31.0      & 1.7                & 5.1              & 2.4          & 40.8          & 26.7  & 19.6           \\
3                    & -w/o Extra Patch Token                        &\textbf{52.7}      &34.6      &31.0       & 1.6      & 4.8              & 2.2          & 40.7          & 26.2   & 19.4            \\
% 3                    & -w/o Spatial Information                    & 50.0     & 34.0     & 30.4      & -                                & -               &          &          &            \\
4                    & -w/o Temporal \& Spatial Expert                             & 48.3     & 32.5     & 29.4      & 1.3     & 3.6               & 1.8          & 37.2          & 24.5  & 18.5            \\
5                    & -w/o Temporal Head                          & 43.6     & 28.3     & 24.8      & 1.1      & 3.4                 & 1.3          &29.4          & 19.3     & 15.7       \\ 
\bottomrule
\end{tabular}

\label{ablation_module}
\end{table*}

% \textsuperscript{{\boldsymbol{$\uparrow$}}}

% \multicolumn{5}{l}{\textit{\textbf{Full-training}}} 

% \multicolumn{5}{l}{\textit{\textbf{Zero-Shot}}}       \\ [1.5pt]

% \vspace{-0.73cm}\hspace{0.8cm}

\smallskip
\noindent\textbf{Effectiveness of the \texttt{<LOC>} Token Interaction.} A key feature of VideoExpert is its decoupling of temporal grounding from text response, enabling different experts to focus on specific tasks while collaborating through a dedicated \texttt{<LOC>} token. This section investigates how the \texttt{<LOC>} token affects model performance. As a comparison, we also report performance variation when VideoExpert employs different interaction strategies between the \texttt{<LOC>} token and T-tokens. As shown in Table~\ref{ablation_loc}, omitting the \texttt{<LOC>} token significantly reduces the model’s grounding ability. Since the Temporal Expert and Spatial Expert process different types of inputs through independent parameters, the absence of the \texttt{<LOC>} token disrupts communication between the two experts, making it difficult for the Temporal Expert to determine when and which events should be localized. This issue is particularly pronounced when grounding tasks require reasoning or involve multiple events. Next, we explore various interaction strategies between the \texttt{<LOC>} token and T-tokens, including direct addition (Add), concatenation (Concat.), and cross-attention (Atten.). The performance differences among these strategies are minimal, which implies that when only the localization of events needs to be indicated, a simple addition of the \texttt{<LOC>} token is sufficient. Therefore, VideoExpert adopts this straightforward addition method for implementation.

% Figure -- 4 and 5
\begin{figure}[!t] %[htbp]是自动排版；[H]固定位置
% \lefting %图片居中
\includegraphics[width = 0.49\textwidth]{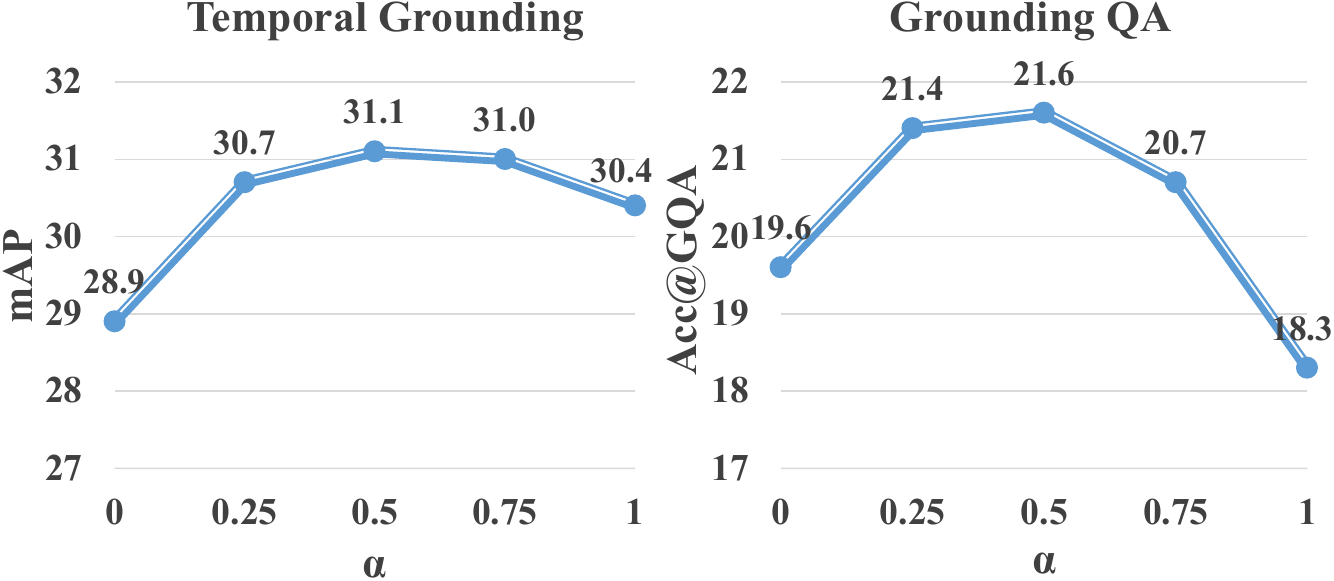} % width = 0.5\textwidth  double_4.pdf
\caption{\textbf{Effectiveness of Different Expert Ranks.} The default LoRA rank in VideoExpert is fixed at $64$ to maintain a constant number of training parameters relative to previous works. The hyperparameter $\alpha$ controls the rank allocation between the two expert components. The rank of the temporal expert is defined as $64 \alpha$, while the spatial expert is given as $64(1 - \alpha)$.}
\label{Rank_alpha} %图片名称和图片标号
\vspace{-1 em}
\end{figure} %结束

\smallskip
\noindent\textbf{Effectiveness of the Components.} In this section, each component is sequentially dropped from the VideoExpert to evaluate the effectiveness of the proposed framework. The results are shown in Table~\ref{ablation_module}. Overall, each component contributes to improving performance. Specifically, the Spatial Compress module is removed in Row 2. In this setting, the constructed S-token is replaced by an equal number of randomly selected patch tokens. In Row 3, all extra patch tokens are completely removed, so the visual input of the model consists solely of the T-token. The results from these two settings show varying effects across different tasks. The impact of the temporal grounding task is minimal. However, for content generation, particularly dense video captioning tasks, the absence of detail-rich inputs leads to a noticeable decline in performance. In Row 4, the Temporal and Spatial Expert are abandoned, and the two expert modules are replaced by a single LoRA module, while the total number of parameters remains unchanged. This change results in a decline across almost all metrics, highlighting the importance of the proposed expert strategy. The Temporal and Spatial Experts each have independent parameter sets. This parameter decoupling design allows for specialized learning within each part without mutual interference. Finally, the temporal head is dropped in Row 5, causing the model to lose its ability to directly generate the timestamp result. As compensation, temporal grounding is converted into a text generation task, following previous work such as~\cite{VTimeLLM}. The final model architecture is degraded to be consistent with the baseline VTimeLLM.

\begin{figure*}[ht]
    \centering
    \includegraphics[width=1\linewidth]{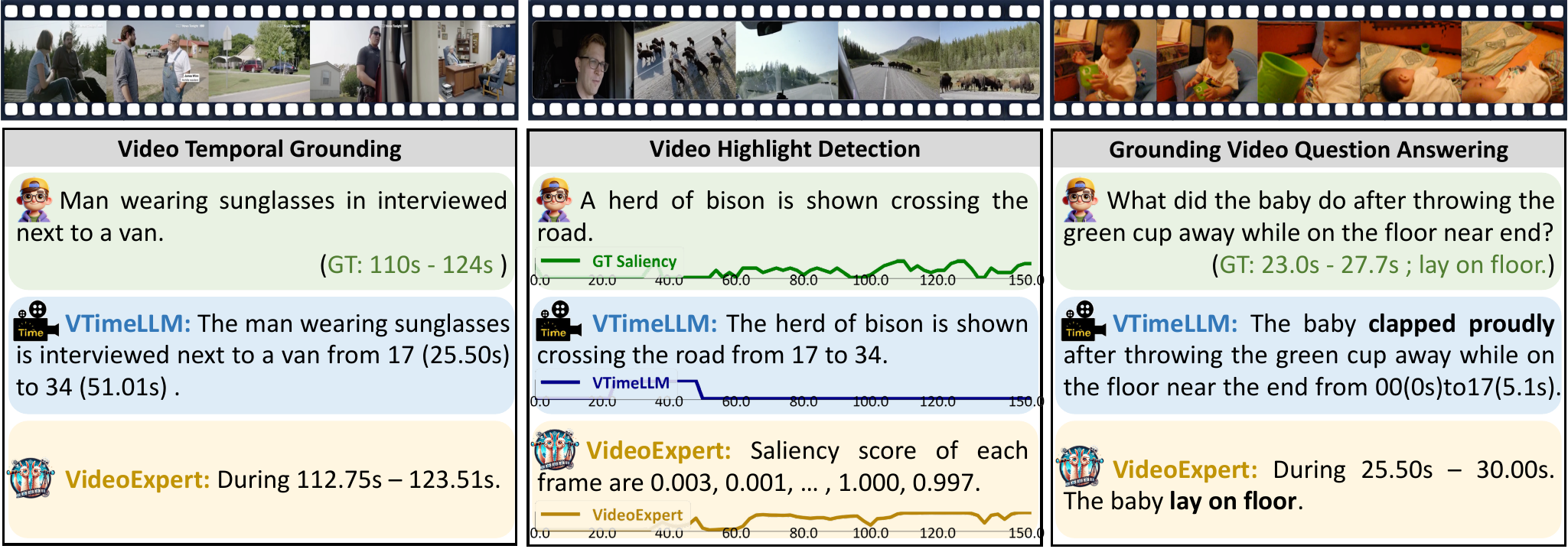}
    \caption{Qualitative results of VideoExpert on Temporal Grounding, Highlight Detection and Grounding QA tasks. }
  \label{fig_QualitativeResult}
  \vspace{-0.5em}
\end{figure*}

\begin{table}[!t]
\centering
\caption{{\upshape Effectiveness of the Spatial Compress module. A total of $n+(u \times w)$ tokens are input into LLM. $n$ represents the number of T-tokens. $u$ denotes the number of GOPs, and $w$ is the number of S-tokens contained in each GOP.}}
\renewcommand{\arraystretch}{1.1}
\begin{tabular}{cl|ccc}
\hline
\multicolumn{1}{c}{\multirow{2}{*}{\textbf{Row}}} & \multicolumn{1}{c|}{\textbf{Input \#tokens}} & \textbf{Inference Speed}      & \textbf{Memory} & \textbf{DVC}  \\
\multicolumn{1}{c}{}                     & \multicolumn{1}{c|}{$n$ + ($u$ $\times$ $w$)}  & Token/s $\uparrow$              & GB $\downarrow$                      & CIDE\textsubscript{r} $\uparrow$ \\ \hline
1                                        & 100 + (0 $\times$ 0)                           & 31.17                & 16.79                   & 4.8      \\ \hline \hline
2                                        & 100 + 4 $\times$ 64                            & 30.61                & 16.79                   & 6.0      \\
3                                        & 100 + 4 $\times$ 128                           & 29.42                & 16.82                   & 6.1     \\
4                                        & 100 + 8 $\times$ 64                            & 29.17                & 16.82                   & 6.4      \\
5                                        & 100 + 8 $\times$ 128                           & 28.39                & 16.96                   & 6.4     \\ \hline
\end{tabular}

\label{training_strategy}
\vspace{-.5em}
\end{table}

%DVC/CIDE\textsubscript{r}
%\begin{tabular}[c]{@{}c@{}}DVC\\ CIDE\textsubscript{r}\end{tabular}

\begin{comment}
    
\begin{tabular}{cl|ccc}
\hline
Row & \multicolumn{1}{c|}{\begin{tabular}[c]{@{}c@{}}Input \#tokens\\ T  + (K * M)\end{tabular}} & Inference Speed (token/s) & Memory used &\begin{tabular}[c]{@{}c@{}}DVC\\ CIDE\textsubscript{r}\end{tabular}    \\ \hline
0 &100 + (0 $\times$ 0)            &   &      &   \\ \hline \hline
% 100$\times$16$\times$16      &    &     &      \\ \hline
1 &100 + 4 $\times$ 64       &     &    &      \\
2 &100 + 4 $\times$ 128      &     &    &      \\ \hline
3 &100 + 8 $\times$ 32       &     &    &      \\
4 &100 + 8 $\times$ 64       &     &    &      \\ \hline
\end{tabular}

\end{comment}

\smallskip
\noindent\textbf{Different Ranks of the Two Experts.} To investigate the contribution of the Temporal and Spatial Expert to VideoExpert, we compared the effects of different rank configurations of the two experts on model performance. In our experiments, the default LoRA rank in VideoExpert is fixed at $64$ to maintain a constant number of training parameters relative to previous works. A hyperparameter $\alpha$ is used to control the rank allocation between the Temporal and Spatial Expert components. Specifically, the rank of the temporal expert is defined as $64 \alpha$, while the spatial expert is given as $64(1 - \alpha)$.

As shown in Figure~\ref{Rank_alpha},  VideoExpert achieves optimal performance at $\alpha=0.5$, where the parameters of the Temporal and Spatial Expert are equal. Deviations from this balanced value, whether towards higher or lower $\alpha$, result in diminished model performance. Furthermore, the combined use of both Temporal and Spatial Expert consistently outperforms the utilization of either strategy in isolation. For example, when $\alpha$ is set to $0$, the temporal expert fails to assist the LLM in perceiving temporal information, leading to poor temporal grounding performance. This finding indicates that both experts are pivotal constituents of VideoExpert to achieve better performance.

\smallskip
\noindent\textbf{Spatial Compress module in VideoExpert.} This section investigates the impact of various settings in our Spatial Compress module. The variable $n$ represents the number of T-tokens, while $(u\times w)$ denotes the total number of S-tokens generated by the Spatial Compress module. Here, $u$ represents the number of GOPs, and $w$ refers to the number of S-tokens within each GOP. Thus, the total number of input tokens to the LLM is computed as $n+(u\times w)$. From Table~\ref{training_strategy}, we observe that increasing the number of S-tokens moderately decreases inference speed, but the drop is minimal. The memory footprint remains stable, fluctuating only slightly across different settings. Furthermore, the CIDE\textsubscript{r} score improves significantly compared to the baseline, demonstrating that the module enhances content quality by extracting and compressing more informative spatial features. Notably, the model benefits further gains from increasing v outweighing those from increasing $w$ in terms of content generation quality. Overall, the additional computational costs and efficiency impacts of using S-tokens are minimal and acceptable given the improvements they bring.

\begin{table}[!t]
\footnotesize
\centering
\caption{{\upshape \small{\textbf{\upshape Video-based Text Generation Benchmarking results.} VideoExpert not only enables accurate temporal localization, but also generally improves video understanding for Video LLMs.}}}
% \vspace{-0.5em}
\renewcommand{\arraystretch}{1.1}
\setlength{\tabcolsep}{4 pt}

\begin{tabular}{lcccccc}
\toprule
\multicolumn{1}{l}{\textbf{Method}} & \textbf{Corr.} & \textbf{Detail} & \textbf{Context} & \textbf{Temp.} & \textbf{Consis.} & \textbf{Mean} \\ \midrule
Video-ChatGPT \cite{VideoChatGPT}                       & 2.40                 & 2.52            & 2.62             & 1.98              & 2.37                 & 2.38             \\
BT-Adapter \cite{btadapter}                          & 2.68                 & 2.69            & 3.27             & 2.34              & 2.46                 & 2.69             \\
VTimeLLM \cite{VTimeLLM}                            & 2.78                 & 3.10            & 3.40             & 2.49              & 2.47                 & 2.85             \\
LLaMA-VID \cite{llamavid}                           & 2.96                 & 3.00            & 3.53             & 2.46              & 2.51                 & 2.89             \\
VideoChat-v2 \cite{videochat2}                        & 3.02                 & 2.88            & 3.51             & 2.66              & 2.81                 & 2.98             \\
% \textcolor{lightgray}{LITA(13B)\cite{LITA}}     & \textcolor{lightgray}{2.94}  & \textcolor{lightgray}{2.98}  & \textcolor{lightgray}{3.43}            & \textcolor{lightgray}{2.68}              & \textcolor{lightgray}{3.19}                 & \textcolor{lightgray}{3.04}             \\
CAT \cite{cat}                                 & 3.08                 & 2.95            & 3.49             & 2.81              & 2.89                 & 3.07             \\
\textbf{Ours}                                & \textbf{3.13}        & \textbf{3.15}   & \textbf{3.61}    & \textbf{2.93}     & \textbf{3.13}        & \textbf{3.19}    \\ \bottomrule
\end{tabular}

\label{Gen_val}
\vspace{-1 em}
\end{table}

\begin{figure}[!tp]
  \centering
  \includegraphics[width=0.95\linewidth]{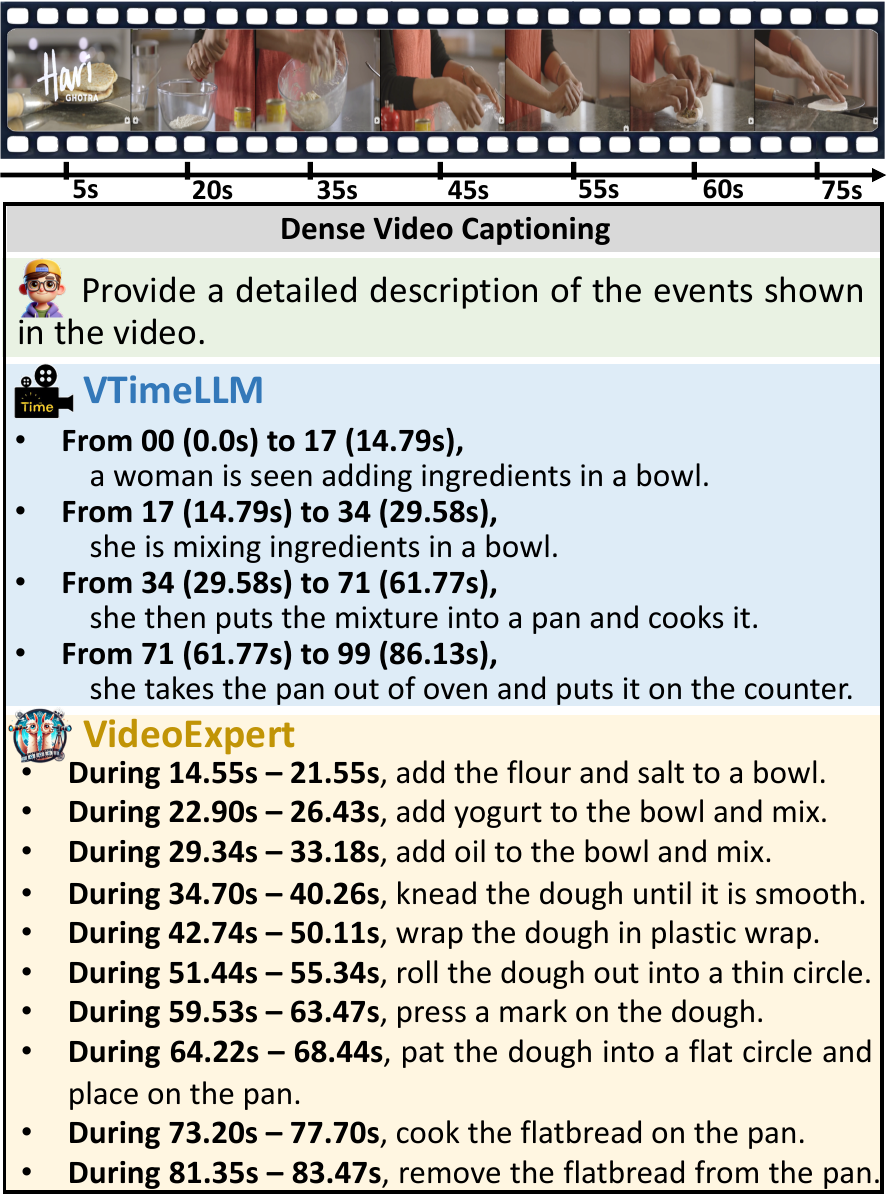}
  \caption{{Qualitative results of VideoExpert on Dense Video Captioning task.}}
  \label{fig_QuantityDVC}
  \vspace{-0.5 cm}
\end{figure}

% \subsection{Video-Based Text Generation Evaluation}
\smallskip
\noindent\textbf{Video-Based Text Generation Evaluation.} In addition to assessing its ablite on temporal-related tasks, we conduct a comprehensive evaluation of VideoExpert's text generation capabilities. Specifically, we use the ``Video-based Text Generation Performance Benchmarking" proposed by Maaz et al~\cite{VideoChatGPT}. This benchmark, built on the ActivityNet-200 dataset~\cite{activity200}, includes a set of videos with rich, detailed captions and human-annotated question-answer pairs. This contrasts with existing video question-answering benchmarks, which typically feature short, concise answers. The questions in this benchmark range from specific inquiries to open-ended ones, enabling a more in-depth evaluation of video understanding. Furthermore, the evaluation leverages GPT-3.5, which scores the model’s responses across five key dimensions: Correctness of Information, Detail Orientation, Contextual Understanding, Temporal Understanding, and Consistency, with scores ranging from 1 to 5. The average scores for each method are then reported to reflect overall performance.

As shown in Table~\ref{Gen_val}, VideoExpert outperforms the competitors across all evaluated fronts, with the most notable improvement observed in temporal understanding. This highlights that our approach excels in temporal-related tasks and enhances overall video comprehension. We attribute these gains to two primary factors. Firstly, the unique architecture of the VideoExpert prevents interference between content understanding and temporal proception, while fostering collaboration across tasks. Secondly, joint training with temporal-related tasks helps the model capture more video content details, ultimately improving its overall comprehension.

\subsection{Qualitative Result}

The qualitative results of VideoExpert are shown in Figures~\ref{fig_QualitativeResult} and~\ref{fig_QuantityDVC}. The yellow blocks indicate the answers predicted by our method, while the blue blocks represent the predictions from VTimeLLM for comparison. Overall, VideoExpert not only localizes events in the video more accurately but also provides detailed content descriptions. In the Highlight Detection task, VideoExpert can assign a saliency score to each frame, rather than returning a time interval. In the Grounding QA task, VideoExpert demonstrates its ability for temporal reasoning, providing correct answers along with the corresponding event timestamps as visual evidence. In the Dense Video Captioning task, VideoExpert lists the specific times when detailed actions occur and explains the content. These are in contrast to the more generic answers by VTimeLLM.

% === IV. Conclusion ========================================
% =================================================================================

%%%%%%%%%%%%%%%%%%%%%%
% Conclusion
%%%%%%%%%%%%%%%%%%%%%

\section{Conclusion}

This paper introduces ‌VideoExpert‌, a video-centric multimodal language model. It is designed to mitigate the language-pattern bias in existing methods for temporal-sensitive video tasks by offloading temporal grounding from text generation. Specifically, VideoExpert decouples these two processes by integrating two parallel expert modules. The temporal expert‌ is dedicated to dynamic modeling and temporal grounding, while the spatial expert‌ focuses on spatial details and instruction following. These two experts operate with independent parameter sets, enabling specialized learning in each part without mutual interference. Through the synergistic integration of both experts‌, VideoExpert effectively mitigates the over-reliance on language patterns in timestamp prediction, and enhances overall performance. Furthermore, a spatial compression module‌ is introduced to optimize efficiency by selectively preserving critical visual information, thereby reducing redundant computations and delivering compact yet detail-rich input for the spatial expert. Experimental results on four datasets across various settings affirm the superior performance of VideoExpert in temporal grounding and content generation. We hope this work can inspire future research to explore the potential of the MLLMs in a series of temporal-sensitive tasks for building robust and resource-efficient models in video-centric AI systems.

\section*{Acknowledgment} 
 % This work was supported by the National Natural Science Foundation of China (Grant No. 62425603) and the Basic Research Program of Jiangsu Province (Grant BK20240011). 
 
We extend our gratitude to Dr. Deng-Ping Fan (Nankai University) for his invaluable guidance and support during my CSC studies.  Dr. Fan provided detailed advice, which contributed to shaping the research topic, experimental design, and the review of this paper.

% or


% ============================================
%\appendices
%Appendix one text goes here %\cite{Roberg2010}.

%\section{}

%\section*{Acknowledgment}

%The authors would like to thank D. Root for the loan of the SWAP. The SWAP that can ONLY be usefull in Boulder...

\ifCLASSOPTIONcaptionsoff
  \newpage
\fi



%\begin{thebibliography}{1}


\bibliographystyle{IEEEtran}
\bibliography{IEEEabrv,Bibliography}

\vfill


\end{document}